\documentclass{article}

\usepackage{microtype}
\usepackage{graphicx}
\usepackage{subfigure}
\usepackage{booktabs} 
\usepackage{natbib}
\usepackage{amssymb}
\usepackage{amsmath}
\usepackage{bm}

\DeclareMathOperator{\argmax}{argmax}

\usepackage[accepted]{icml2018}

\icmltitlerunning{Scalable Natural Gradient Langevin Dynamics in Practice}

\begin{document}
    
\twocolumn[
\icmltitle{Scalable Natural Gradient Langevin Dynamics in Practice}



\icmlsetsymbol{equal}{*}

\begin{icmlauthorlist}
\icmlauthor{Henri Palacci}{col}
\icmlauthor{Henry Hess }{col}
\end{icmlauthorlist}
\icmlaffiliation{col}{Department of Biomedical Engineering, Columbia University}

\icmlcorrespondingauthor{Henri Palacci}{hp2393@columbia.edu}

\icmlkeywords{Machine Learning, ICML}

\vskip 0.3in
]


\printAffiliationsAndNotice{}  

\begin{abstract}
Stochastic Gradient Langevin Dynamics (SGLD) is a sampling scheme for Bayesian modeling adapted to large datasets and models. SGLD relies on the injection of Gaussian Noise at each step of a Stochastic Gradient Descent (SGD) update. In this scheme, every component in the noise vector is independent and has the same scale, whereas the parameters we seek to estimate exhibit strong variations in scale and significant correlation structures, leading to poor convergence and mixing times. We compare different preconditioning approaches to the normalization of the noise vector and benchmark these approaches on the following criteria: 1) mixing times of the multivariate parameter vector, 2) regularizing effect on small dataset where it is easy to overfit, 3) covariate shift detection and 4) resistance to adversarial examples.     
\end{abstract}

\section{Introduction}                                                                                                                                                
Deep Learning is moving into fields for which errors are potentially lethal, such as self-driving cars, healthcare, and biomedical imaging.
For these applications, being able to estimate errors is essential.
Bayesian methods provide a way to expand scalar predictions to full posterior probabilities \cite{Gelman2014-um}.            
Stochastic Gradient Langevin Dynamics (SGLD), is one of the solutions to the issue of probabilistic modeling on large datasets. Gaussian noise is added to the SGD updates \cite{Welling2011-zy}. It was proposed to pre-condition the Gaussian noise with a diagonal matrix to adapt to the changing curvature of the parameter space \cite{Li2016-lq}. Using a full preconditioning matrix corresponding to the metric tensor of the parameter space was previously proposed \cite{Girolami_Mark2011-pj}, but the computation of this tensor is impossible for large-scale neural networks.  It was further proposed to use the Kronecker-factored block diagonal approximation of this tensor, first introduced in \cite{Martens2015-ix} and \cite{Grosse2016-ab} as the preconditioning tensor for the Langevin noise \cite{Nado2018-bh}. Fixed learning rate vanilla gradient descent also introduces noise in the learning process. Hence, fixed learning rate SGD can also be seen as a variant on the same method \cite{Mandt2017-jn}.         
                                                                             
In this paper, we conduct a comparison of all these approaches in a practical setting with a fixed hyperparameter optimization budget. We compare these approaches using traditional Markov Chain Monte Carlo (MCMC) diagnostic tools, but will also evaluate the: performance of models in recognizing data points that are not in the sample distribution, the reduction of overfitting in small data settings,  and the robustness to adversarial attacks. We find that Langevin approaches, with a reasonable computing budget for hyperparameter tuning, do not improve overfitting or help with adversarial attacks. However, we do find a significant improvement in the detection of out-of-sample data using Langevin methods.                                  
    
\section{Related Work}                                                                                                                                           
SGLD was introduced in \cite{Welling2011-zy} and was further refined using a diagonal preconditioning matrix (pSGLD) in \cite{Li2016-lq}. The natural gradient method was introduced by \cite{Amari1998-fs}. Girolami and Calderhead proposed to extend the natural gradient method to neural networks in \cite{Girolami_Mark2011-pj}, and a practical application to probability simplices was presented in \cite{Patterson2013-ju}. Finally, the interpretation of fixed rate SGD (FSGD) as a Bayesian approximation was shown in \cite{Mandt2017-jn}. The Kronecker-Factored block-diagonal approximation of the inverse Fisher information matrix was presented for dense layers in \cite{Martens2015-pp}, then extended to convolutional layers in \cite{Grosse2016-ab}. This was used as a preconditioning matrix in SGLD (KSGLD) for smaller scale experiments in \cite{Nado2018-bh}.         

\section{Preliminaries}                                                                                                                                               
\subsection{Probabilistic Neural Networks}                                                                                                                            
We consider a supervised learning problem, where we have data $\mathbf{x}_1, ..., \mathbf{x}_n \in \mathbb{R}^p$, and labels $y_1, ..., y_n$ drawn from a distribution $\mathcal{P}$. Our goal is to approximate the distribution $p(y|\mathbf{x})$ by empirical risk minimization of a family of distributions parametrized by a vector $\bm{\theta}$.

In the non-probabilistic setting, this is done by defining an appropriate loss function $\mathcal{L}(y_i | \mathbf{x}_i; \bm{\theta_i})$ and minimizing it with respect to $\bm{\theta}$. Optionally, a regularizing term $\mathcal{R}(\bm{\theta})$ is added to the minimization problem which can therefore be written as: $\bm{\hat{\theta}} = \argmax \sum_i -\mathcal{L}(y_i, x_i; \bm{\theta}) + \mathcal{R}(\bm{\theta}) $. This can be understood as the MAP estimate of the probabilistic model $p(\bm{\theta}|\mathbf{x}) =  p(\bm{\theta})\prod_i p(y_i, x_i| \bm{\theta})$, where $p(\bm{\theta}|\mathbf{x})$ is the posterior probability of the parameters, $\ln p(\bm{\theta})=  \mathcal{R}(\bm{\theta}) $ is the log-prior, and $\ln p(y_i, \mathbf{x}_i|\bm{\theta})=\mathcal{L}(y_i, x_i; \bm{\theta}) $ is the log-likelihood.                                                                                  
         
\subsection{Stochastic Gradient Langevin Dynamics} \label{SGLDsection}
                                                                                                                                                                      
The workhorse algorithm for loss minimization for neural networks is mini-batch stochastic gradient descent (SGD).
The data $\mathbf{x}_1, ... \mathbf{x}_n$ is grouped into mini batches $B_1, ..., B_j, ...$ of size $J$ such that $(\mathbf{x}_1, ... \mathbf{x}_J) \in B_1, (\mathbf{x}_{J+1}, ..., \mathbf{x}_{2J}) \in B_2, ...$      
                                                                                                                                                                                                                                
Stochastic Gradient Langevin Dynamics (SGLD) \cite{Welling2011-zy} updates modifies SGD by adding Gaussian noise at each update step: $\Delta\bm{\theta}_{t} = \lambda_t \nabla_{\theta} \left( \log{p(\bm{\theta})} + \sum_j \log{p(B_j, \bm{\theta}})\right) + \bm{\epsilon}$, where $\bm{\epsilon} \sim \mathcal{N}(0, \lambda_t \mathbf{I})$. 

\subsection{Riemaniann Manifold Langevin Dynamics}                                                                                                                    
                                                                                                                                                                      
The space formed by the parameters of a probability distribution is a Riemaniann manifold \cite{Amari1998-fs}. Its Riemaniann metric is the Fisher information matrix. This means that the parameter space is curved, and that a local measure of curvature is the Fisher information matrix:    $F(\theta) = \mathbb{E}\left[ \partial_\theta p(y|x;\theta) \partial_\theta p(y|x;\theta)^T\right] $.  Riemaniann Manifold Langevin Dynamics \cite{Marceau-Caron2017-qt}  preconditions the SGD update with the inverse of the Fisher information matrix: $\Delta\bm{\theta}_{t} = F^{-1}\lambda_t \nabla_{\theta} \left( \log{p(\bm{\theta})} + \sum_j \log{p(B_j, \bm{\theta}})\right) + F^{-1}\bm{\epsilon}$. Unfortunately, the computation of the inverse Fisher information matrix is impossible in very high dimensional spaces.                                          
                                                                                                                                                                      
\subsection{Kronecker-Factored Approximate Curvature}                                                                                                                 
                                                                                                                                                                      
The Kronecker-Factored Appoximate Curvature (KFAC) is a compact and efficiently invertible block-diagonal approximation of the Fisher information matrix proposed in \cite{Martens2015-ix} for dense layers of neural networks and in \cite{Grosse2016-ab} for convolutional layers. Each block corresponds to a layer of the neural network, hence this approximation correctly takes into account within-layer geometric structure. Each layer $i$'s activations $a_i$ can be computed from the previous layer's activations by a matrix product $s_i = \mathbf{W}a_{i-1}$. A non-linear activation function $\phi$ such that $a_i = \phi(s_i)$ is applied. The K-FAC approximation can then be written using the Kronecker product $\otimes$: $\widetilde{F} = \text{diag} \left( A_1 \otimes G_1, ..., A_i \otimes G_i, ..., A_l \otimes G_l\right)$, where $A_i = \mathbb{E} \left[a_i a_i^T \right]$ is the estimated covariance matrix of activations for layer $i$, and $G_i = \mathbb{E} \left[g_i g_i^T \right]$ where $g_i = \nabla_s \mathcal{L}(y, x; \theta)$. We can invert the Kronecker product of two matrices by $ (A \otimes B)^{-1} = A^{-1} \otimes B^{-1} $, and can therefore compute the approximate inverse Fisher information matrix as $\widetilde{F}^{-1} = \text{diag} \left(  \{A_i^{-1} \otimes G_i^{-1}\}_{i=1...l}\right)$.

\subsection{Scalable Natural Gradient Langevin Dynamics}                                                                                                              
                                                                                                                                                                      
To implement a tractable preconditioning inverse matrix, \cite{Li2016-lq} used a diagonal preconditioning matrix rescaling the noise by the inverse of its estimated variance (pSGLD). Although this improves on SGLD, it still neglects the off-diagonal terms of the metric. A quasi-diagonal approximation was proposed in \cite{Marceau-Caron2017-qt}. Here, we follow the results presented in \cite{Nado2018-bh} and use the K-FAC approximation to the inverse Fisher information matrix as our preconditioning matrix:                                                                                                  
                                                                                                                                                                      
\begin{equation}                                                                                                                                                      
    \label{KSGD}                                                                                                                                                      
    \Delta\bm{\theta}_{t} = \widetilde{F}^{-1}\lambda_t \nabla_{\theta} \left( \log{p(\bm{\theta})} + \sum_j \log{p(B_j, \bm{\theta}})\right) + \widetilde{F}^{-1}\bm{\epsilon}                                                                                                                                                             
\end{equation}                                                                                                                                                        
                                                                                                                                                                      
Notice that when changing preconditioning matrices in practice, it is unclear if any improvement in convergence of the algorithms comes from preconditioning the gradient term above, or from preconditioning the noise. It is one of the questions that we aim to answer with our experiments.                                             
                                                                                                                                                                      
\subsection{Fixed Learning Rate Stochastic Gradient Descent}                                                                                                          
                                                                                                                                                                      
It has been suggested that traditional SGD, using a decreasing schedule for the learning rate and early stopping performs Bayesian updates \cite{Mandt2017-jn}. The noise introduced by the variability in the data also prevents the posterior from collapsing to the MAP.                                          
                                  
\section{Experiments}                                                                   
In order for the model comparisons to be fair, we used the same neural network architecture for all experiments: two convolutional layers with 32 and 64 layers and max-pooling, followed by one dense layer with 1024 units. All nonlinearities are ReLU. The hyperparameter optimization was run using grid search, and the computational time for hyperparameter optimization was limited to 5 times that of the standard SGD algorithm for all other algorithms. Batch size for all experiments was 512.      
                                                                                                                                                                      
Note that we did not apply the preconditioning matrix to the gradient term. It is otherwise impossible to tell if the performance improvements come from better gradient updates in the initial, non-Langevin part of training or from the improvement of the latter, steady-state part of training. Our SGD updates are therefore:         
\begin{equation}                                                                                                                                                      
    \Delta\bm{\theta}_{t} = \lambda_t \nabla_{\theta} \left( \log{p(\bm{\theta})} + \sum_j \log{p(B_j, \bm{\theta}})\right) + \widetilde{G}\bm{\epsilon}              
\end{equation}                                                                                                                                                        
                                                                                                                                                                      
Where $G=\mathbf{0}$ for SGD, $G=\mathbf{I}$ for SGLD, $G$ is the diagonal RMSprop matrix for pSGD, $G=\widetilde{F}^{-1}$ for KSGD, and $\lambda_t =\lambda$ for fixed learning rate SGD (FSGD).                   
\subsection{Test Set Accuracy}

\begin{figure}[ht]
\vskip 0.1in
\begin{center}
\centerline{\includegraphics[width=\columnwidth]{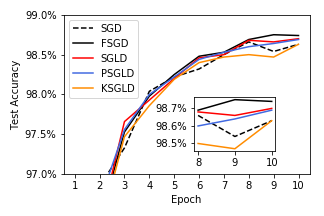}}
\caption{Test set accuracy over ten epochs on the MNIST dataset. SGD: Stochastic Gradient Descent, SGLD: Stochastic Gradient Langevin Dynamics, pSGLD: preconditioned SGLD, KSGLD: K-FAC preconditioned SGLD, FSGD: Fixed rate SGD. Inset: Test set accuracy for the last three epochs.}
\label{accuracy}
\end{center}
\vskip -0.2in
\end{figure}

We first compare the test set accuracy for all methods on 10 epochs of training on the MNIST dataset \cite{LeCun2010-lv}. The results are shown in Figure \ref{accuracy}; accuracies for all models are very close and, for a reasonable hyperparameter tuning budget, Bayesian averaging of models does not seem to improve test set accuracy. 

For the SGLD, pSGLD, and KSGLD methods, the results were very sensitive to the learning rate schedule decrease and most of the hyperparameter optimization computation time was spent on the optimizing it. A longer time spent optimizing the learning rate schedule improved the test rate accuracies slightly. 

\subsection{Mixing Performance}

We approximate \cite{Vats2015-pp} and estimate the effective sample size as: $\text{mESS} = n\left(\frac{\lvert\Lambda\lvert}{\lvert\Sigma\lvert} \right)^{1/p}$,
with $n$ the number of samples in the chain, $p$ the parameter space dimension, $\lvert\Sigma\lvert$ is the covariance matrix of the chain, and $\lvert\Lambda\lvert$ the covariance of matrix of samples. We approximate this by the diagonal approximation of both these matrices, where the ratio of the diagonal terms $\text{ess}_i$ is computed as follows $\text{ess}_i = \frac{n}{1+2\sum_k \rho_k}$, where $\rho_k$ is the autocorrelation at lag $k$ truncated to the highest lag with positive autocorrelation \cite{Gelman2014-um}.

\begin{figure}[ht]
\vskip 0.1in
\begin{center}
\centerline{\includegraphics[width=\columnwidth]{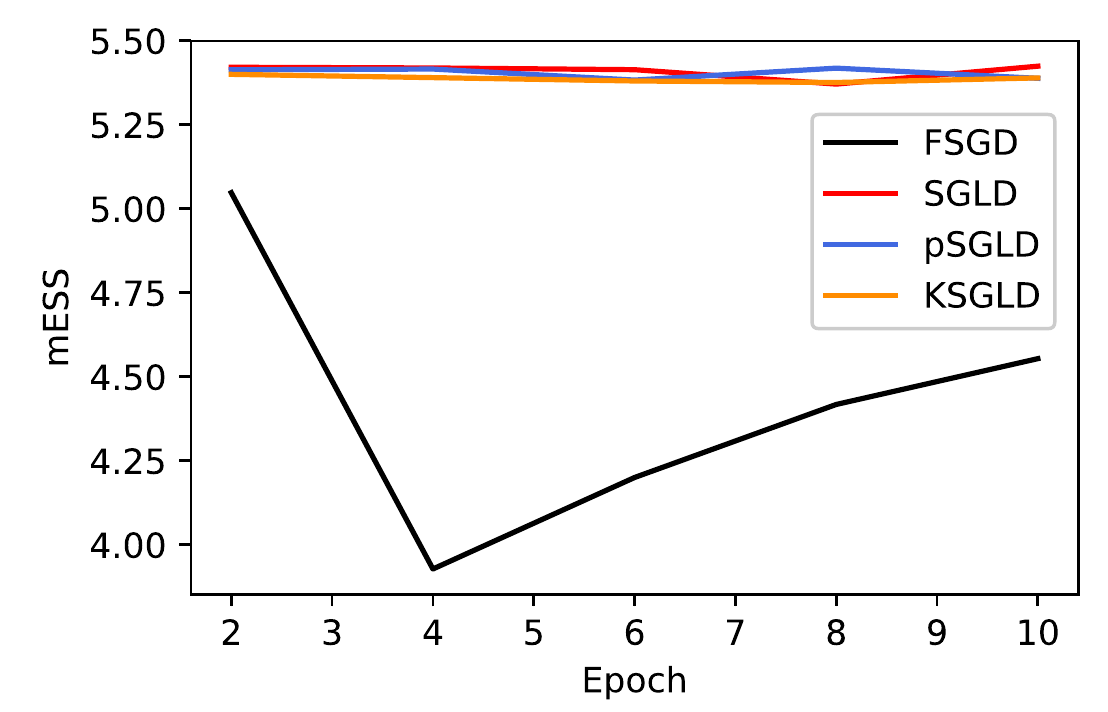}}
\caption{Multivariate Sample Size over epochs for each model over 10 epochs of MNIST training.}
\label{ess}
\end{center}
\vskip -0.2in
\end{figure}

The results, shown in Figure \ref{ess}, all indicate that the MCMC chain mixes poorly in practical settings. Further inspection of the traces shows that almost none of the parameters are stationary. Increasing the run length, or increasing the rate of decrease of the step $\lambda_t$, did not improve the aspect of the traces or the effective sample size. These results are consistent with the theoretical analysis of \cite{Betancourt2015-wo}, who shows that data subsampling is incompatible with any HMC procedure. This is also consistent with \cite{Vollmer2015-en} highlighting the problem of stopping while step sizes are still finite.

\subsection{Reduction of Overfitting}

To test the implicit regularization for the Langevin dynamic models, we truncated the MNIST train set to 5,000 examples (from 60,000). The CNN overfits to the small training set promptly, resulting in decreases in the test set accuracy. 

\begin{figure}[ht]
\vskip 0.1in
\begin{center}
\centerline{\includegraphics[width=\columnwidth]{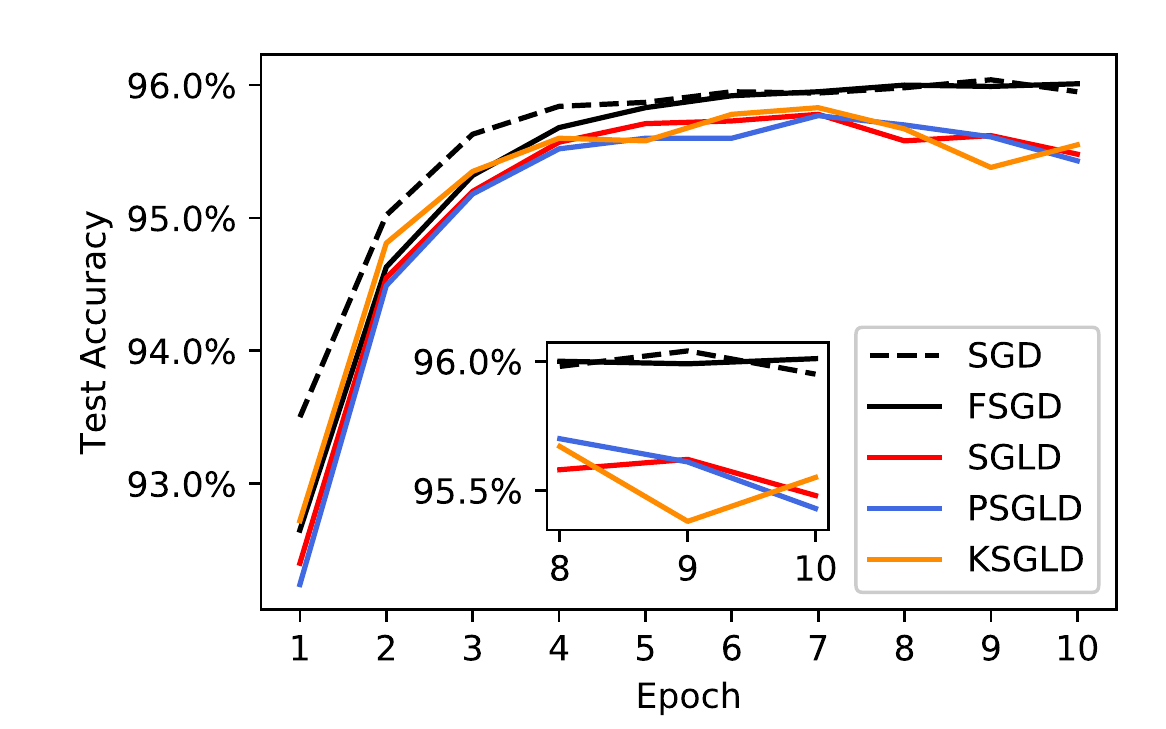}}
\caption{Test set accuracy for all models on ten epochs of training on the reduced MNIST dataset, smallMNIST}
\label{smallmnist}
\end{center}
\vskip -0.2in
\end{figure}

The results, shown in Figure \ref{smallmnist}, show that the dynamic models underperform SGD on smallMNIST. The only dynamic Bayesian method that matches SGD is SGDA. We hypothesize that adding Gaussian noise on such a small amount of data dramatically deteriorates the initial period of convergence, thus forcing the dynamic Langevin methods to settle for the Langevin period in a local minimum of the loss surface.

\subsection{Resistance to Adversarial Attacks}

Adversarial attacks are imperceptible modifications to data that cause a model to fail \cite{Goodfellow2014-wm}. We compute adversarial modifications to the test set using the Fast Gradient Sign Method from \cite{Goodfellow2014-wm}. It has previously been shown in \cite{Rawat2017-fy} that other Bayesian deep learning methods such as Monte Carlo dropout,\cite{Gal2015-lm}, Bayes by Backprop \cite{Blundell2015-tt}, matrix variational gaussian \cite{Louizos2016-ty}, and probabilistic backpropagation \cite{Hernandez-Lobato2015-oc} are vulnerable to adversarial attacks. Our results, presented in Table \ref{adversarialTable}, show that all Langevin dynamic methods also fail to detect adversarial attacks.

\begin{table}[ht]
\caption{Classification accuracies for naive Bayes and flexible
Bayes on various data sets.}
\label{adversarialTable}
\vskip 0.15in
\begin{center}
\begin{small}
\begin{sc}
\begin{tabular}{lcccr}
\toprule
Model & Test  & Accuracy on \\
 & Accuracy & Adversarial Examples \\
\midrule
SGD    & 96.0 & 2.9 \\
FSGD & 96.5 & 2.0 \\
SGLD    & 97.2 & 1.8 \\
pSGLD    & 97.1 & 1.9  \\
KSGLD     & 97.0 & 2.0\\
\bottomrule
\end{tabular}
\end{sc}
\end{small}
\end{center}
\vskip -0.1in
\end{table}

\subsection{Detection of Out of Sample Examples}

We assess the epistemic uncertainty inherent in our Bayesian deep neural networks by training it on MNIST but evaluating the network on a completely different dataset, notMNIST \cite{Bulatov_undated-hj}. The notMNIST dataset is similar in format to the MNIST dataset, but consists of letters from different fonts.

\begin{figure}[ht]
\vskip 0.1in
\begin{center}
\centerline{\includegraphics[width=\columnwidth]{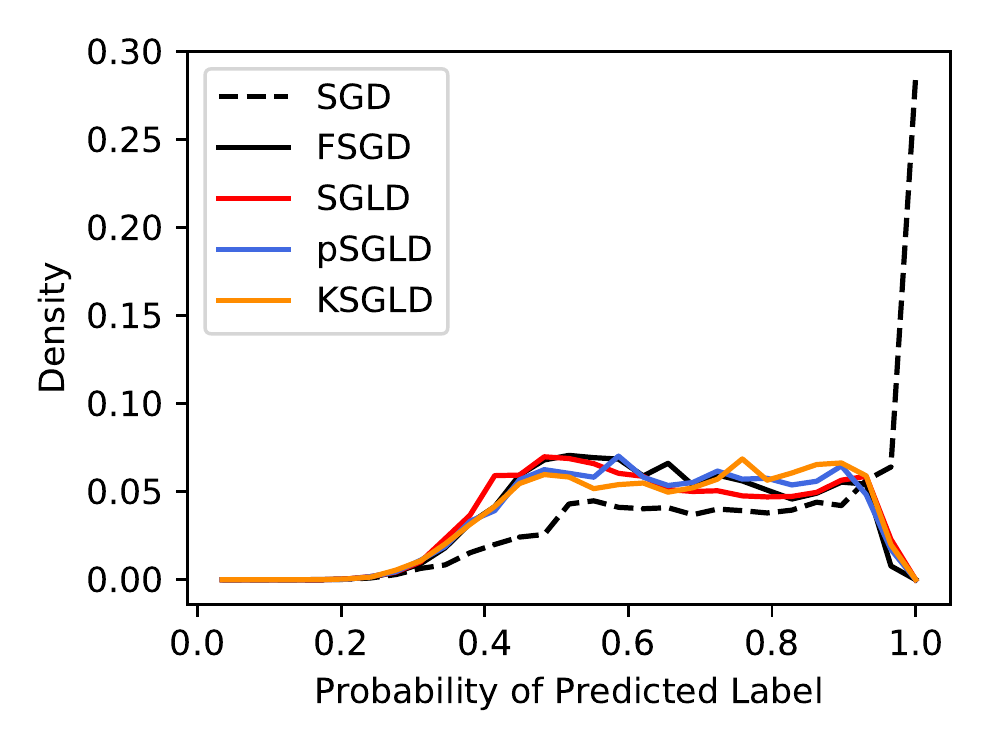}}
\caption{Probability distribution for the most likely class on the notMNIST dataset for all models trained on the MNIST dataset. }
\label{notmnist}
\end{center}
\vskip -0.2in
\end{figure}

We expect a network trained on MNIST to give relatively low class probabilities when given examples from the notMNIST dataset. Figure \ref{notmnist} shows the distribution of the highest probability for each example. Vanilla SGD gives very confident predictions for this dataset, whereas all other methods present a similar distribution of uncertainties. This suggests that Langevin dynamics and fixed learning rate SGD are a relatively straightforward way to detect covariate shift in practical classification tasks.

\section{Discussion}

Langevin Stochastic Dynamics provide a scalable way to compute Bayesian posteriors on deep neural network architectures. The noise in stochastic gradient Langevin dynamics is not isotropic due to the geometry of the parameter space. To render the Gaussian noise isotropic, diagonal \cite{Li2016-xf}, quasi-diagonal \cite{Marceau-Caron2017-qt}, and block-diagonal \cite{Martens2015-ix} approximations have been used. These preconditioning matrices have been proven to work very well as preconditioners for the gradient term, but their use as preconditioners for the Gaussian term in SGLD is subject to significant convergence issues, especially in the transition from the learning phase, where the mini-batch noise dominates. 

By contrast, leveraging the mini-batch noise by a constant learning rate to prevent posterior collapse seems to work just as well as the Langevin methods for the experiments described above. This could suggest that the 'data noise' is already appropriately scaled to the manifold structure of the parameter space. This will be the subject of future research.

In practice, our experiments suggest to use Bayesian averaging with a fixed learning rate; this doesn't require any modification to the standard training workflows used by practitioners, and provides implicit protection against covariate shift.

\section*{Acknowledgements}

This work was supported by the ARO grant "Thermodynamics of Statistical Learning", PI: H. Hess, ARO W911-NF-17-1-0107.

\bibliographystyle{icml2018}
\bibliography{paperpile}

\begin{thebibliography}{24}
\providecommand{\natexlab}[1]{#1}
\providecommand{\url}[1]{\texttt{#1}}
\expandafter\ifx\csname urlstyle\endcsname\relax
  \providecommand{\doi}[1]{doi: #1}\else
  \providecommand{\doi}{doi: \begingroup \urlstyle{rm}\Url}\fi

\bibitem[Amari(1998)]{Amari1998-fs}
Amari, Shun-Ichi.
\newblock Natural gradient works efficiently in learning.
\newblock \emph{Neural Comput.}, 10\penalty0 (2):\penalty0 251--276, February
  1998.

\bibitem[Betancourt(2015)]{Betancourt2015-wo}
Betancourt, Michael.
\newblock The fundamental incompatibility of scalable hamiltonian monte carlo
  and naive data subsampling.
\newblock In \emph{International Conference on Machine Learning}, pp.\
  533--540. jmlr.org, June 2015.

\bibitem[Blundell et~al.(2015)Blundell, Cornebise, Kavukcuoglu, and
  Wierstra]{Blundell2015-tt}
Blundell, Charles, Cornebise, Julien, Kavukcuoglu, Koray, and Wierstra, Daan.
\newblock Weight uncertainty in neural networks.
\newblock May 2015.

\bibitem[Bulatov()]{Bulatov_undated-hj}
Bulatov, Yaroslav.
\newblock {notMNIST} dataset.
\newblock http://yaroslavvb.blogspot.com/2011/09/notmnist-dataset.html.
\newblock Accessed: 2018-4-24.

\bibitem[Gal \& Ghahramani(2015)Gal and Ghahramani]{Gal2015-lm}
Gal, Yarin and Ghahramani, Zoubin.
\newblock Dropout as a bayesian approximation: Representing model uncertainty
  in deep learning.
\newblock June 2015.

\bibitem[Gelman et~al.(2014)Gelman, Carlin, Stern, Dunson, Vehtari, and
  Rubin]{Gelman2014-um}
Gelman, Andrew, Carlin, John~B, Stern, Hal~S, Dunson, David~B, Vehtari, Aki,
  and Rubin, Donald~B.
\newblock \emph{Bayesian data analysis}, volume~2.
\newblock CRC press Boca Raton, FL, 2014.

\bibitem[{Girolami Mark} \& {Calderhead Ben}(2011){Girolami Mark} and
  {Calderhead Ben}]{Girolami_Mark2011-pj}
{Girolami Mark} and {Calderhead Ben}.
\newblock Riemann manifold langevin and hamiltonian monte carlo methods.
\newblock \emph{J. R. Stat. Soc. Series B Stat. Methodol.}, 73\penalty0
  (2):\penalty0 123--214, March 2011.

\bibitem[Goodfellow et~al.(2014)Goodfellow, Shlens, and
  Szegedy]{Goodfellow2014-wm}
Goodfellow, Ian~J, Shlens, Jonathon, and Szegedy, Christian.
\newblock Explaining and harnessing adversarial examples.
\newblock December 2014.

\bibitem[Grosse \& Martens(2016)Grosse and Martens]{Grosse2016-ab}
Grosse, Roger and Martens, James.
\newblock A kronecker-factored approximate fisher matrix for convolution
  layers.
\newblock \emph{arXiv:1602.01407 [cs, stat]}, February 2016.

\bibitem[Hern{\'a}ndez-Lobato \& Adams(2015)Hern{\'a}ndez-Lobato and
  Adams]{Hernandez-Lobato2015-oc}
Hern{\'a}ndez-Lobato, Jos{\'e}~Miguel and Adams, Ryan~P.
\newblock Probabilistic backpropagation for scalable learning of bayesian
  neural networks.
\newblock February 2015.

\bibitem[LeCun et~al.(2010)LeCun, Cortes, and Burges]{LeCun2010-lv}
LeCun, Yann, Cortes, Corinna, and Burges, C~J.
\newblock {MNIST} handwritten digit database.
\newblock \emph{AT\&T Labs [Online]. Available: http://yann. lecun.
  com/exdb/mnist}, 2, 2010.

\bibitem[Li et~al.(2016{\natexlab{a}})Li, Chen, Carlson, and Carin]{Li2016-lq}
Li, Chunyuan, Chen, Changyou, Carlson, David~E, and Carin, Lawrence.
\newblock Preconditioned stochastic gradient langevin dynamics for deep neural
  networks.
\newblock In \emph{{AAAI}}, volume~2, pp.\ ~4, 2016{\natexlab{a}}.

\bibitem[Li et~al.(2016{\natexlab{b}})Li, Chen, Carlson, and Carin]{Li2016-xf}
Li, Chunyuan, Chen, Changyou, Carlson, David~E, and Carin, Lawrence.
\newblock Preconditioned stochastic gradient langevin dynamics for deep neural
  networks.
\newblock In \emph{{AAAI}}, volume~2, pp.\ ~4, 2016{\natexlab{b}}.

\bibitem[Louizos \& Welling(2016)Louizos and Welling]{Louizos2016-ty}
Louizos, Christos and Welling, Max.
\newblock Structured and efficient variational deep learning with matrix
  gaussian posteriors.
\newblock March 2016.

\bibitem[Mandt et~al.(2017)Mandt, Hoffman, and Blei]{Mandt2017-jn}
Mandt, Stephan, Hoffman, Matthew~D, and Blei, David~M.
\newblock Stochastic gradient descent as approximate bayesian inference.
\newblock \emph{arXiv:1704.04289 [cs, stat]}, April 2017.

\bibitem[Marceau-Caron \& Ollivier(2017)Marceau-Caron and
  Ollivier]{Marceau-Caron2017-qt}
Marceau-Caron, Ga{\'e}tan and Ollivier, Yann.
\newblock Natural langevin dynamics for neural networks.
\newblock \emph{arXiv:1712.01076 [cs, stat]}, December 2017.

\bibitem[Martens \& Grosse(2015{\natexlab{a}})Martens and
  Grosse]{Martens2015-ix}
Martens, James and Grosse, Roger.
\newblock Optimizing neural networks with kronecker-factored approximate
  curvature.
\newblock In \emph{International Conference on Machine Learning}, pp.\
  2408--2417, 2015{\natexlab{a}}.

\bibitem[Martens \& Grosse(2015{\natexlab{b}})Martens and
  Grosse]{Martens2015-pp}
Martens, James and Grosse, Roger.
\newblock Optimizing neural networks with kronecker-factored approximate
  curvature.
\newblock March 2015{\natexlab{b}}.

\bibitem[Nado et~al.(2018)Nado, Snoek, Grosse, Duvenaud, Xu, and
  Martens]{Nado2018-bh}
Nado, Zachary, Snoek, Jasper, Grosse, Roger, Duvenaud, David, Xu, Bowen, and
  Martens, James.
\newblock {STOCHASTIC} {GRADIENT} {LANGEVIN} {DYNAMICS} {THAT} {EXPLOIT}
  {NEURAL} {NETWORK} {STRUCTURE}.
\newblock February 2018.

\bibitem[Patterson \& Teh(2013)Patterson and Teh]{Patterson2013-ju}
Patterson, Sam and Teh, Yee~Whye.
\newblock Stochastic gradient riemannian langevin dynamics on the probability
  simplex.
\newblock In \emph{Advances in Neural Information Processing Systems}, pp.\
  3102--3110, 2013.

\bibitem[Rawat et~al.(2017)Rawat, Wistuba, and Nicolae]{Rawat2017-fy}
Rawat, Ambrish, Wistuba, Martin, and Nicolae, Maria-Irina.
\newblock Adversarial phenomenon in the eyes of bayesian deep learning.
\newblock November 2017.

\bibitem[Vats et~al.(2015)Vats, Flegal, and Jones]{Vats2015-pp}
Vats, Dootika, Flegal, James~M, and Jones, Galin~L.
\newblock Multivariate output analysis for markov chain monte carlo.
\newblock \emph{arXiv:1512.07713 [math, stat]}, December 2015.

\bibitem[Vollmer et~al.(2015)Vollmer, Zygalakis, and Teh]{Vollmer2015-en}
Vollmer, Sebastian~J, Zygalakis, Konstantinos~C, and Teh, Yee~W.
\newblock (non-) asymptotic properties of stochastic gradient langevin
  dynamics.
\newblock January 2015.

\bibitem[Welling \& Teh(2011)Welling and Teh]{Welling2011-zy}
Welling, Max and Teh, Yee~W.
\newblock Bayesian learning via stochastic gradient langevin dynamics.
\newblock In \emph{Proceedings of the 28th International Conference on Machine
  Learning ({ICML-11})}, pp.\  681--688, 2011.

\end{thebibliography}

\end{document}